\documentclass{article}
\usepackage{spconf,amsmath,graphicx}
\usepackage{enumitem}
\usepackage{bbm}
\usepackage{hyperref}

\title{Cross-attention watermarking of large language models}

\name{Folco Bertini Baldassini$^{1}$ \qquad Huy H. Nguyen$^{2}$ \qquad Ching-Chung Chang$^{2}$ \qquad Isao Echizen$^{2, 3}$\thanks{Code available at \href{https://gitlab.com/folbaeni/linguistic-watermark}{https://gitlab.com/folbaeni/linguistic-watermark}\\This work was partially supported by JSPS KAKENHI Grants JP18H04120, JP20K23355, JP21H04907, and JP21K18023, and by JST CREST Grants JPMJCR18A6 and JPMJCR20D3, Japan.}}

\address{\small{$^{1}$Sorbonne University, France  \ \ \ \ \ \ \ \  $^{2}$National Institute of Informatics, Japan \ \ \ \ \ \ \ \ $^{3}$The University of Tokyo, Japan}}

\newcommand*{\flamingo}{gated cross attention layer }
\newcommand*{\lastlayer}{decoder layer substitution }

\begin{document}
\ninept
\maketitle
\begin{abstract}
A new approach to linguistic watermarking of language models is presented in which information is imperceptibly inserted into the output text while preserving its readability and original meaning. A cross-attention mechanism is used to embed watermarks in the text during inference. Two methods using cross-attention are presented that minimize the effect of watermarking on the performance of a pretrained model. Exploration of different training strategies for optimizing the watermarking and of the challenges and implications of applying this approach in real-world scenarios clarified the tradeoff between watermark robustness and text quality. Watermark selection substantially affects the generated output for high entropy sentences. This proactive watermarking approach has potential application in future model development.
\end{abstract}

\vspace{-0.5em}
\section{Introduction}
\vspace{-0.5em}
Linguistic watermarking refers to the insertion of particular unnoticeable information into a text document while preserving its readability, intended meaning, and ability to withstand noise~\cite{8268096}. Interest in such watermarking is growing due to the widespread emergence of AI-generated text, which poses risks in various sectors, including online influence campaigns, AI exploitation of authorship, spamming, harassment, malware facilitation, and social engineering~\cite{goldstein2023generative, crothers2022machine}. Text watermarking has various applications, including the tagging of AI-generated text. It can be used to protect intellectual property, detect leaks, and verify the source. In addition, it aids in complying with regulations that mandate identification of AI-generated texts in countries that are implementing such requirements~\cite{helberger2023chatgpt}. Lastly, watermarking can assist language model developers in ensuring that model output is distinguishable to prevent model collapse during training on generated text~\cite{shumailov2023curse}.

As the output of large language models (LLMs) approaches the quality of human-generated text, it becomes more difficult to distinguish between them. As LLMs improve, increase in number, and become easier to fine-tune, the existing post-hoc detectors for AI-generated text are becoming more unreliable in correctly identifying such content. The defenses often fail to generalize to real-world scenarios~\cite{sadasivan2023can} and often fail in adversarial settings~\cite{pu2022deepfake}. Intentionally modifying models to embed watermarks in the output text is a promising solution to this challenge. Unfortunately, the implementation of a useful watermark often involves substantial modifications to the distribution of text generated by the template, potentially degrading its quality~\cite{cryptoeprint:2023/763}.

Our focus is on blind watermarking, which enables verification based solely on bits extracted from the watermarked text~\cite{8268096}. They can be used to identify information, copyrights, digital signatures, or any other element that can establish the source or integrity of the generated text. Watermarking ensures accurate content attribution, assists in plagiarism detection, and prevents the malicious manipulation of information. With this approach, nouns, verbs, adjectives, and other grammatical elements are modified without changing the text's intended meaning~\cite{8268096}. The exploitation of semantic information in text content has shown promise in improving the robustness and generalization performance of deepfake text detection systems~\cite{pu2022deepfake}.

However, adversaries have the goal of damaging a watermark with minimal changes to the text. To counter this, a highly robust watermark is needed that can withstand being heavily modified and still remain effective. Such a watermark should discourage attackers because it would make their efforts futile. Furthermore, the generated text should be of high quality to ensure that the model remains viable. To summarize, a watermark should be able to seamlessly integrate and authenticate itself within the original text while making minimal alterations to the original content. In addition, watermarks should be imperceptible and indistinguishable from the original content even when subjected to repeated adaptive queries. Furthermore, watermarks should be robust against simple attempts to remove them and remain extremely unlikely to produce false positives.

Although numerous studies of watermarking using machine learning have been reported, few have specifically addressed the LLM domain and exploited the underlying LLM architecture to produce watermarked output in real time during inference. We have addressed this shortcoming. This work makes four key contributions:
\begin{itemize}[leftmargin=*,noitemsep,topsep=0pt]
    \item A watermarking layer has been designed that incorporates a cross-attention mechanism into the language model. This approach exploits the LLM architecture to produce robust watermarks during inference with a minimal increase in parameter count.
    \item An explainable framework for watermark verification has been devised, and the challenges and implications of applying the proposed watermarking approach in real-world scenarios have been identified.
    \item Two methods using cross-attention are presented that minimize the effect of watermarking on the performance of a pretrained model.
    \item Training strategies that enhance the performances and the robustness of the proposed watermarking technique have been developed.
\end{itemize}
\vspace{-0.5em}
\vspace{-0.5em}

\section{Related Work}
\vspace{-0.5em}
Natural language watermarking methods~\cite{atallah2001natural,atallah2002natural} initially used syntactic trees and large prime numbers as secret keys to insert watermarks into natural language text. However, changes in language models made these methods obsolete. An alternate approach~\cite{topkara2006hiding} uses word sense ambiguity to embed watermarks by replacing words with labelled synonyms. More advanced techniques~\cite{xiang2018reversible} include a reversible watermarking method that combines arithmetic coding and synonym substitution operations. Machine learning models have also been applied to text watermarking, initially using long short-term memory (LSTM) models~\cite{fang2017generating} to select tokens carrying specific bits and thereby ensuring that the generated watermarked text followed the desired distribution. The original data can be recovered in a deterministic manner by mapping the tokens back to their corresponding bit blocks.

The advent of transformers led to the development of methods that use sequence incremental watermarking with an infill model~\cite{yang2022tracing} or a neural paraphraser~\cite{qiang2023natural} whose candidates are selected by semantic relevance and target bit. Siamese models like Siamese-BERT (SBERT)~\cite{reimers2019sentence} enable the determination of semantic similarity among sentences. There are methods that utilize a pretrained Word2Vec model~\cite{munyer2023deeptextmark, mikolov2013efficient} or an infill model like BERT (Bidirectional Encoder Representations from Transformers)~\cite{yoo2023robust, devlin2018bert} for generating candidate replacement words. SBERT is used to evaluate the quality of each proposed sentence. The proposed sentence with the greatest similarity to the original sentence becomes the watermarked sentence. A pretrained BERT model is utilized for watermark detection using a binary classifier.

Abdelnabi and Fritz~\cite{abdelnabi2021adversarial} introduced a method that uses an encoder-decoder transformer to rewrite the text and embed a binary vector as a watermark. The model consists of an embedder that uses an encoder-decoder transformer, a bidirectional extractor to reconstruct the binary vector, and a bidirectional discriminator that preserves language statistics. SBERT is used to incorporate semantic loss, and a custom LSTM model is used for grammatical correction loss. Unfortunately, these models necessitate identical embeddings, which may be less discernible with contemporary tokenizers, such as SentencePiece~\cite{kudo2018sentencepiece}.

Kirchenbauer et al.~\cite{kirchenbauer2023watermark} presented a comprehensive strategy for watermarking text generated by LLMs. Their approach involves categorizing the tokens as either 'green' or 'red' on the basis of a key with the aim of getting the model to produce more 'green' words. This is achieved by assigning more weight to the 'green' words before applying the softmax operation. Statistical tests are used to detect the presence of a watermark. This method utilizes the functioning of transformers and applies watermarks during inference at minimal cost. The watermark is directly inserted during text generation. However, token selection for promotion or avoidance is randomized rather than using a contextually informed approach. Moreover, this approach provides only watermark detection functionalities; it is not able to integrate a message into the produced text. These limitations motivated us to devise novel structures that enable watermarking within the model directly.

\vspace{-0.5em}
\section{Proposed Methods}
\vspace{-0.5em}
The proposed methods use a two-part system comprising a watermark embedder for watermarked text generation and a watermark extractor for watermark verification. During text generation, the embedder processes a textual prompt and binary vector watermark to produce watermarked text as output. It does this by using a pretrained LLM enhanced with a watermark module and then fine-tuned. The purpose of this enhancement is to preserve the quality of the generated text while adding the watermark.

The extractor module analyzes the input sentence and produces a binary vector representing the extracted watermark. If the input text does not include a watermark, the resulting vector is a sequence of random bits, making it impossible to verify. However, a modified watermarked document must still yield an extracted watermark that partly corresponds to the authentication bits or displays visible indications of identifying information. The process of information retrieval using symbiotic embedding and recovery yields fewer false positives. This approach is reliable even when handling hybrid human-machine texts. The retrieval procedure for the watermark has adaptable capabilities even if only a part of the watermark is preserved.

\vspace{-0.5em}
\subsection{Models}
\begin{figure}
      \centering
      \includegraphics[width=0.88\linewidth]{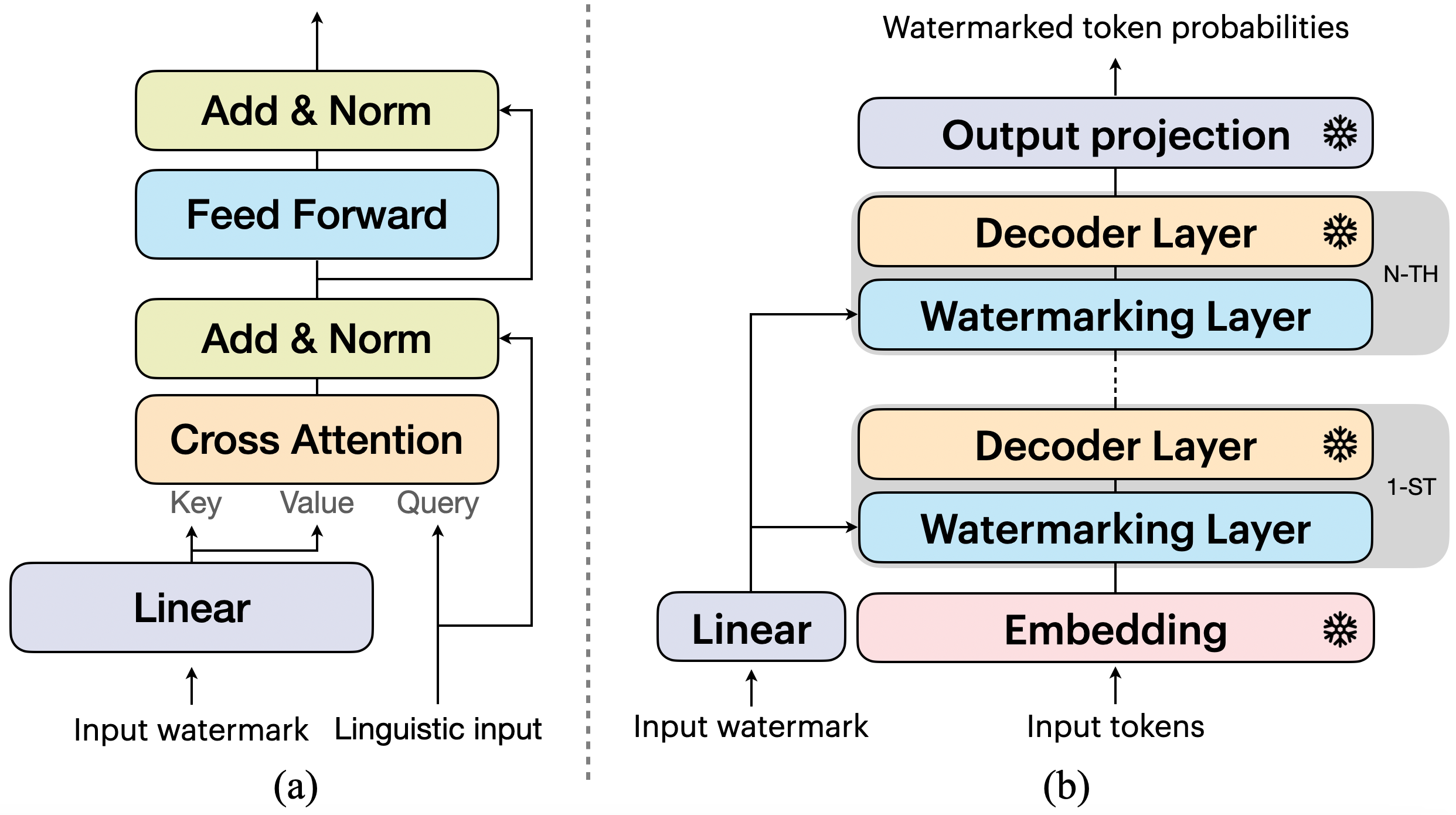}
      \vspace{-1em}
      \caption{(a) Cross-attention mechanism for embedding watermark; linguistic input can be embeddings or self-attention. (b) Watermark layers, each with cross-attention and feedforward block, are place between pretrained decoder layers.}
      \label{fig:xattention_flamingo}
      \vspace{-1.5em}
\end{figure}

Our study of cross-attention draws inspiration from work on multimodal models that use it to incorporate images~\cite{alayrac2022flamingo,chen2022visualgpt}. Chen et al.~\cite{chen2022visualgpt} included cross-attention in pretrained layers, similar to the encoder-decoder attention mechanism suggested in the original transformer~\cite{vaswani2017attention}. Alayrac et al.~\cite{alayrac2022flamingo} introduced a layer with gated cross-attention and feedforward connections between pretrained layers. These approaches focus on integrating images into textual models, which necessitates the handling of a great amount of visual information. In our scenario, only a few pertinent bits per sentence are conveyed, so a lighter and more parameter-efficient solution is needed. We use a linear layer to integrate the watermark into the model dimension through cross-attention (Figure~\ref{fig:xattention_flamingo}(a)). Similar to multimodal models, the watermark embedding serves as the key and value, while the linguistic input forms the query.

The \textbf{\flamingo} approach is based on the work of Alayrac et al.~\cite{alayrac2022flamingo}. The linear layer is structured as follows: $w$ is the embedded watermark and $\alpha$ and $\beta$ are parameters initialized to zero to maintain the model output and ensure stability.
\vspace{-.5em}
\[out = \text{LayerNorm}\big(y + \text{FeedForward}(y) * tanh(\beta)\big)\]
\vspace{-1.5em}
\[y = \text{LayerNorm}\big(x + \text{Attention}(q=w, kv=x) * tanh(\alpha)\big)\]
We place gated layers before the decoder layers (Figure~\ref{fig:xattention_flamingo}(b)). We avoid introducing unnecessary parameters by limiting their use to only the last N layers since the watermark has fewer bits than an image. During training, we freeze the entire base model and optimize only the cross-attention components.

The \textbf{\lastlayer} approach involves taking a pretrained model, either decoder-only or encoder-decoder, and replacing the last decoder layer with one that includes self-attention, cross-attention, and feedforward connections. This new layer keeps the original model's positional embeddings, biases, etc. In the initial phase of training, we freeze the entire base model and only unfreeze the output projection layer at convergence, thereby modifying the original model. This step is necessary to avoid a drop in performance, which enables the model to further reduce watermarking loss. The \lastlayer approach is a cost-effective strategy during inference and even if it were made open source, it would still require fine-tuning to eliminate the watermarking layer.

The watermark extractor is based on a bidirectional transformer model, which leverages its ability to capture contextual information from both preceding and succeeding tokens. Following transformer output, a pooling mechanism aggregates the information, creating a compact representation of the extracted features. These pooled features are passed through a multi-layer perceptron, and a binary vector is extracted. 

The use of a pretrained model (in our case \textit{stabilityai/stablelm-base-alpha-7b} from HuggingFace) requires the placement of constraints on the hyperparameters of the new decoder layer such as the hidden and feedforward dimensions and the bit precision. This also applies to the extractor since it relies on the same embeddings. For a model with 7 billion parameters, the \flamingo approach adds a total of 720 million (3 layers) parameters, whereas the \lastlayer approach adds 330 million parameters, and the extractor has 1 billion.

\vspace{-0.5em}
\subsection{Unambiguous Text Identification}
During watermark verification, the text is examined to establish whether it has been watermarked, i.e., generated by our model. This can be achieved through the implementation of two methods, message embedding and key checking, which have the advantages of content insertion and robustness, respectively.

Message embedding transforms the provided message into a binary vector, which acts as a watermark. The watermark is divided into multiple vectors that match the bit capacity of a single sentence. During the generation phase, these segmented tensors are used sequentially to encode individual sentences. Recurrent bits are incorporated to enhance detection capabilities. During the verification stage, these bits are extracted from each sequence and then concatenated to reconstruct the original vector. If the text does not have a watermark, the bits are random. If the text has a watermark, the resulting vector reflects the original message. If the watermarked text is modified to remove the watermark, or in the case of hybrid human-machine composition, remnants of the original message fragments remain, providing evidence of their origin from our model. The extractor uses a sigmoid activation function to provide confidence levels for each bit. This is crucial when dealing with adversaries who attempt to remove the watermark while preserving the original text. By identifying confident bits as pivots, we can use heuristics, depending on how the message is translated into bits, to search for uncertain bits. This enables message reconstruction even in the presence of corruption.

Alternatively, if the focus is on robustness, the watermark can act as a key. For ease of detection, the same binary vector can be used for each sentence. We would encode the same vector in each sentence, so whenever we extract one, it indicates that the text is watermarked. This can be achieved following the approach of Abdelnabi and Fritz~\cite{abdelnabi2021adversarial}. The null hypothesis $(H_0)$ attributes observed matching bits to chance. This means that the behavior of matching bits (random variable $X$) adheres to a binomial distribution with $n$ as the bit count, $k$ as the matching count, and $0.5$ as the success probability. The p-value, quantifying the likelihood of $k$ or more matches under $H_0$, is
\vspace{-1.5em}\[\Pr(X > k | H_0) = \sum_{i=k}^{n} \binom{n}{i}  0.5^n\vspace{-0.2em}\]
A watermark is verified if its p-value is below threshold $T$, indicating that the occurrence of this specific sequence by mere chance is highly improbable. We thereby obtain a degree of confidence for each sentence. The results can be represented in a graphical interface, with color highlighting used to represent the likelihood of each sentence being watermarked. We can thus obtain valuable insights beyond simple binary classification. This approach is advantageous for hybrid texts and scenarios in which the text has been modified to remove the watermark.

\vspace{-0.5em}
\subsection{Enhancing Robustness and Training Methodology}
\label{training}
The training pipeline consists of several key steps. Initially, the input tokens, in our case obtained from the OpenWebText dataset~\cite{Gokaslan2019OpenWeb}, along with a randomly generated watermark of eight bits per sentence, are fed into the watermarker. The watermark is incorporated into the process through a cross-attention mechanism. The resulting output is then utilized to compute the cross-entropy loss, which is measured against the tokens shifted by one position. Subsequently, the Gumbel-Softmax method, as proposed by Jang et al.~\cite{jang2016categorical}, is used to calculate the one-hot encodings. These encodings are multiplied by the weights of the embedding layers. Finally, the transformed data are passed through the remaining stages of the extractor, and the resulting output is evaluated using binary cross entropy against the original watermark.

\begin{figure}
    \centering
    \includegraphics[width=.8\linewidth]{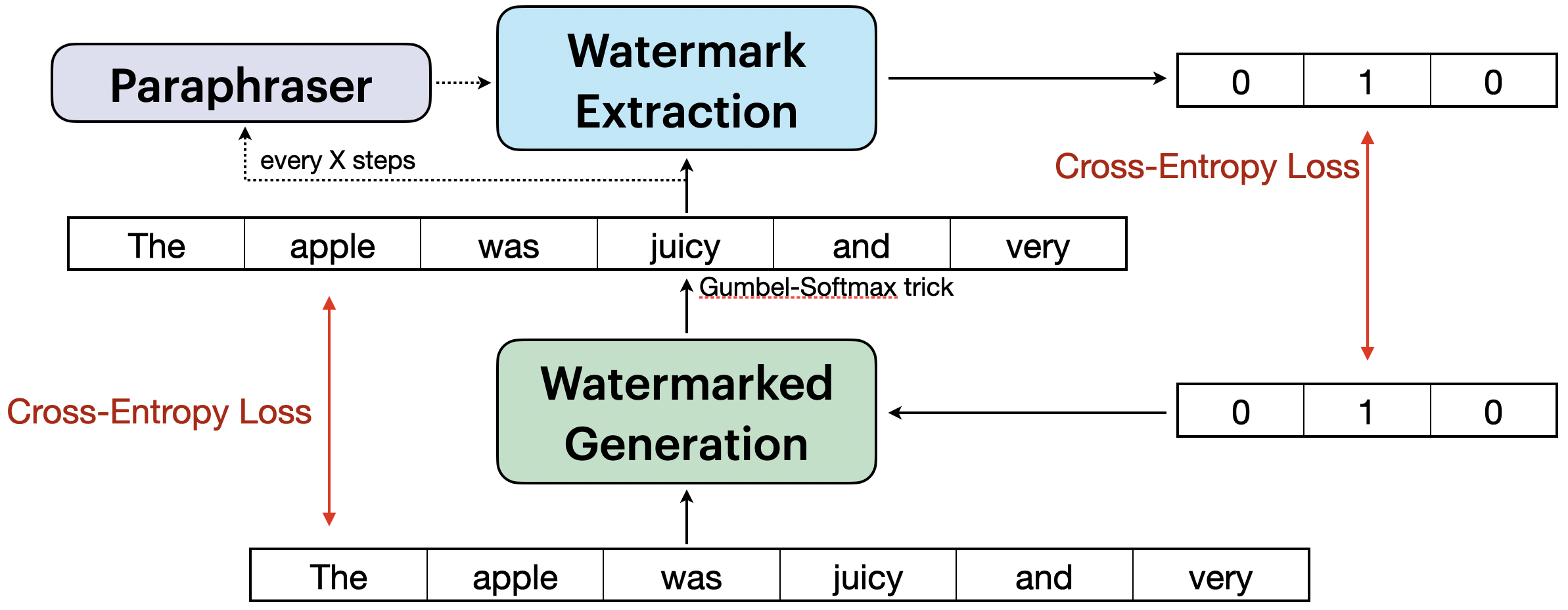}
    \vspace{-1em}
    \caption{High level representation of the training scheme}
    \label{fig:training}
    \vspace{-1.5em}
\end{figure}

To improve the robustness of the embedder, we should simulate an attack on the generated text prior to its use by the extractor.  However, operating at the textual level becomes impractical as conversion between tokens and text disrupts the computational graph. Consequently, in order to improve the robustness of the embedder, a simulated attack which is able to be backpropagated is required. We propose adding Gaussian noise into the extractor after the embedding layer, with the aim of imitating synonym substitution on random tokens for a portion of the sentences. 
\vspace{-.5em}\[
    \Tilde{E}_s^t = 
\begin{cases}
    E_s^t & \text{if } s\leq|S|*p_{\text{sent}}\\
    E_s^t + \text{Bern}(p_{\text{word}}) * \mathcal{N}(0, \sigma^2)              & \text{otherwise}
\end{cases}\vspace{-.5em}
\]
In practice, we sum the embeddings with Bernoulli-masked noise:
\vspace{-.5em}\small{\[
\Tilde{E} = E + \mathcal{N}_{(S,T,E)}(0, \sigma^2) \odot \text{Bern}_{(S,)}(p_{\text{sent}}) \odot \text{Bern}_{(S,T)}(p_{\text{word}})\vspace{-.5em}
\]}
where S, T, E, and p represent the sentences, tokens, embeddings, and probabilities of modification, respectively. This approach ensures backpropagation and facilitates improvement of the embedder's robustness.

Furthermore, we train intependently the extractor with watermarked text as input, which is then subjected to a paraphrasing model (\textit{tuner007\slash pegasus\_paraphrase} from Huggingface). Subsequently, the loss is computed using the original watermark applied during the input phase. This enhances the robustness of our system against machine learning-based automatic attacks.

Regarding attacks, we use a baseline involving the application of synonym substitution on various tokens using Word2Vec~\cite{mikolov2013efficient} as well as the random addition, deletion, and substitution of tokens, each with distinct probabilities\label{baseline}. It is important to note that these modifications degrade the quality of the text, rendering it suboptimal. Undertaking a complete rewrite of the text would require substantial effort and result in minimal modifications to the content, thereby making it challenging to establish the threshold at which the watermark should still persist. Another baseline is evaluating the model's robustness against a paraphraser simulating a common attack that any potential adversary could easily carry out using ChatGPT at a minimal cost.

\vspace{-0.5em}
\section{Findings}
\textbf{Models: } Comparing our approach to others presents challenges as they either focus on detecting watermark presence~\cite{kirchenbauer2023watermark} or on rewriting text~\cite{abdelnabi2021adversarial} while, our method aims to reconstruct the original watermark and seamlessly integrate it during text generation.  Therefore the performance of the cross-attention is compared with that of the adversarial watermarking transformer (AWT) model in Table \ref{tab:baseline}. We adapted the training pipeline~\cite{abdelnabi2021adversarial} by incorporating the watermark with cross-attention in the final layer instead of summing to encoder memory, obtaining slightly better results.

\begin{figure}
    \centering
    \includegraphics[width=.8\linewidth]{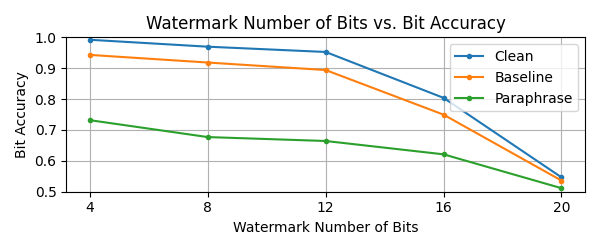}
    \vspace{-1em}
    \caption{Watermark reconstruction bit precision with \lastlayer approach under various attacks.}
    \label{fig:bit}
    \vspace{-1em}
\end{figure}

For both proposed models, the overall text performance was far from competitive. Nevertheless, we obtained reliable watermarking reconstruction performance with legible output.
The \textbf{\flamingo} approach affected the text quality less since we did not modify the original model. However, this constraint poses challenges during training because failure to slightly affect the original model output can lead to total collapse. We were able to achieve 70\% bit accuracy while maintaining text quality. The \textbf{\lastlayer} approach produced more promising outcomes; unfortunately, it considerably alters the original model and requires considerable fine-tuning before it can be put into practice. It attained a considerably higher level of bit precision and capacity (Fig. \ref{fig:bit}); however, the level was inversely proportional to the quality of the text. It initially demonstrated satisfactory text quality but subsequently experienced pronounced deterioration. In particular, with regards to greedy sampling, this deterioration became evident after 64–128 tokens. Although a sampling strategy mitigates this phenomenon, the model remains unreliable for long outputs in its current form. More research is required to address this problem, particularly as models become capable of processing more tokens.

\begin{table}[]
    \centering
    \begin{tabular}{l|l|l}
         & Text Cross Entropy & Watermark Cross Entropy\\ \hline
    AWT  & 2.96 & 0.12     \\
    Ours & \textbf{2.37} & \textbf{0.11}    
    \end{tabular}
    \vspace{-.5em}
    \caption{Training losses on test set with AWT and our method.}
    \vspace{-1em}
    \label{tab:baseline}
\end{table}

\textbf{Augmentation Techniques:} Both noise injection and paraphrasing have been shown to increase bit accuracy (Table~\ref{tab:robustness}). Since both apply to the extractor, they do not have a negative effect on text quality. Noise injection is cost-effective, supporting its integration. In contrast, integrating paraphrasing can substantially increase the computational load during training. Despite the prevalence of paraphraser attacks in contemporary scenarios, it is necessary to explore the benefits of augmentation. Furthermore, the extractor could be trained later over attack examples without modifying the model.

\begin{table}[]
    \centering
    \begin{tabular}{c|c|c|c}
         Augmentation technique & Clean & Baseline & Paraphrase \\\hline
         None               & 0.986             & 0.943 & 0.672\\
         Noise              & \textbf{0.991}    & \textbf{0.953} & 0.688 \\
         Paraphrase         & 0.982             & 0.940 & 0.764 \\
         Paraphrase+Noise   & 0.973             & 0.947 & \textbf{0.774} \\
    \end{tabular}
    \vspace{-1em}
    \caption{Watermark  reconstruction bit precision with \lastlayer approach if text is not modified or attacked using either baseline or paraphrase approach.}
    \label{tab:robustness}
    \vspace{-1.5em}
\end{table}

\textbf{Challenges: } Our study highlights a challenging tradeoff between the capability to create robust watermarks and the quality of the generated text. Achieving high watermark robustness typically comes at the expense of text quality, as exemplified by the straightforward approach of substituting words with specific synonyms, which are easily detectable, leading to poor quality output lacking contextual coherence. Our experiments confirmed that striking the right balance between these two objectives is crucial.

\textbf{Sentence entropy:} When different watermarks are used with identical prompts, the output is highly variable but still syntactically correct and coherent with the prompt. With low entropy, i.e., the first few words strongly dictate the following words, the output remains consistent regardless of the watermark input. Nevertheless, we noticed more pronounced effects of the watermark on the topic of the generated text when using open-ended prompts. For instance, the prompt \textit{"The man was accused of plowing into a group of…"} with various watermarks results in outputs related to protesters, people, capitalists, workers, or even innocent horses (deterministic sampling). However, this apparently does not impede the model's ability to provide rational responses. Interestingly, the fixed nature of a very low entropy sentence makes it unlikely to contain a watermark, which is a notable discovery.

\section{Conclusion}
We have presented an innovative approach to linguistic watermarking for language models. Specific information is imperceptibly embedded into text during inference while preserving readability and intended meaning. The two proposed methods using cross-attention minimize the effect of watermarking on the performance of the pretrained model. Despite a lower overall performance of the LLMs, the results are promising. A particularly intriguing technique is the substitution of the last layer. The integration of multimodal methods with efficient strategies~\cite{hu2022lora} could enhance the potential of this watermarking technique.
We have also introduced two techniques—adding noise and paraphrasing—that improve the resilience of the system, inexpensively and against paraphraser attacks, respectively. We have also introduced two ways of authenticating watermarks, message embedding and use of a fixed watermark vector, each with its own strengths and limitations.

This study sheds light on the potential of watermarking techniques to shape the themes of the generated content, especially with open-ended prompts, raising questions for AI alignment. A key insight is the existence of a tradeoff between watermark robustness and text quality. Achieving a high degree of watermark robustness often means compromising the quality of the generated text, while preserving quality can challenge the resilience of the watermark. Finding the optimal balance between these two objectives remains a critical area for further investigation and improvement. In essence, embedding watermarks directly into large language models has the potential to balance output quality and cost-effectiveness. This proactive approach of integrating watermarks directly during model development has the potential to simplify the watermarking process and ensure widespread adoption.

\newpage
\bibliographystyle{IEEEbib}
\bibliography{references}
\end{document}